%% file: Arxiv.tex
\documentclass[conference,onecolumn]{IEEEtran}

\IEEEoverridecommandlockouts
\usepackage{cite}
\usepackage{amsmath,amssymb,amsfonts}
\usepackage{graphicx}
\usepackage{textcomp}
\usepackage{xcolor}
\def\BibTeX{{\rm B\kern-.05em{\sc i\kern-.025em b}\kern-.08em
		T\kern-.1667em\lower.7ex\hbox{E}\kern-.125emX}}

\usepackage{graphicx}
\usepackage{subcaption}
\usepackage{algorithm}
\usepackage{algpseudocode}
\usepackage{enumitem}
\usepackage{multirow}
\usepackage{tabularx}
\usepackage{threeparttable}
\usepackage{color, xspace}
\usepackage{titlesec}
\usepackage{ragged2e}
\usepackage{lastpage}
\usepackage{fancyhdr}
\usepackage{setspace}
\usepackage{array}
\usepackage{caption}
\usepackage{tabularx}
\usepackage{float}
\usepackage{amssymb}
\usepackage{booktabs}
\usepackage{graphicx}
\usepackage{listings}
\usepackage{booktabs}
\usepackage{amsmath}
\usepackage{mathrsfs}

\ifCLASSINFOpdf
\else
\fi

\begin{document}

\title{Cumulative Distribution Function based General Temporal Point Processes}

\author{\IEEEauthorblockN{Maolin Wang}
	\IEEEauthorblockA{\textit{City University of Hong Kong} \\
		Morin.wang@my.cityu.edu.hk}\\
		
	\IEEEauthorblockN{Ruocheng Guo}
	\IEEEauthorblockA{\textit{ByteDance Research, London} \\
		rguo.asu@gmail.com}\\
		
	\IEEEauthorblockN{Yiqi Wang}
	\IEEEauthorblockA{\textit{Michigan State University} \\
		wangy206@msu.edu}\\
	
	\and
	
	\IEEEauthorblockN{Yu Pan}
	\IEEEauthorblockA{\textit{HIT, Shenzhen} \\
		iperryuu@gmail.com}\\
		
	\IEEEauthorblockN{Xiangyu Zhao$^\ast$}
 \thanks{$^{\ast}$ Xiangyu Zhao is the corresponding author.}
	\IEEEauthorblockA{\textit{City University of Hong Kong} \\
		xianzhao@cityu.edu.hk}\\
		
	\IEEEauthorblockN{Zitao Liu }
	\IEEEauthorblockA{\textit{Jinan University} \\
		zitao.jerry.liu@gmail.com}
	
	\and
		
	\IEEEauthorblockN{Zenglin Xu}
	\IEEEauthorblockA{\textit{HIT, Shenzhen} \\
		zenglin@gmail.com}\\

	\IEEEauthorblockN{Wanyu	Wang}
	\IEEEauthorblockA{\textit{City University of Hong Kong} \\
		wanyuwang4-c@my.cityu.edu.hk}\\

	\IEEEauthorblockN{Langming Liu}
	\IEEEauthorblockA{\textit{City University of Hong Kong} \\
		langmiliu2-c@my.cityu.edu.hk}
}

\maketitle

\begin{abstract}
Temporal Point Processes (TPPs) hold a pivotal role in modeling event sequences across diverse domains, including social networking and e-commerce, and have significantly contributed to the advancement of recommendation systems and information retrieval strategies. Through the analysis of events such as user interactions and transactions, TPPs offer valuable insights into behavioral patterns, facilitating the prediction of future trends. However, accurately forecasting future events remains a formidable challenge due to the intricate nature of these patterns.
The integration of Neural Networks with TPPs has ushered in the development of advanced deep TPP models. While these models excel at processing complex and nonlinear temporal data, they encounter limitations in modeling intensity functions, grapple with computational complexities in integral computations, and struggle to capture long-range temporal dependencies effectively.
In this study, we introduce the CuFun model, representing a novel approach to TPPs that revolves around the Cumulative Distribution Function (CDF). CuFun stands out by uniquely employing a monotonic neural network for CDF representation, utilizing past events as a scaling factor. This innovation significantly bolsters the model's adaptability and precision across a wide range of data scenarios. Our approach addresses several critical issues inherent in traditional TPP modeling: it simplifies log-likelihood calculations, extends applicability beyond predefined density function forms, and adeptly captures long-range temporal patterns.
Our contributions encompass the introduction of a pioneering CDF-based TPP model, the development of a methodology for incorporating past event information into future event prediction, and empirical validation of CuFun's effectiveness through extensive experimentation on synthetic and real-world datasets. The CuFun model, with its profound focus on CDFs, marks a paradigm shift in TPP modeling, offering enhanced adaptability, computational efficiency, and versatile applicability. This positions CuFun as a robust choice for real-world applications where the understanding and forecasting of extended temporal sequences hold paramount importance.
\end{abstract}

\IEEEpeerreviewmaketitle

\input{Content/Introduction.tex}
\input{Content/Proposed_Model.tex}

\input{Content/methods}

\input{Content/Experiment.tex}

\input{Content/Related_Work.tex}

\input{Content/Conclusion.tex}

\bibliographystyle{IEEEtran} 
\bibliography{sample-base}
\end{document}

%% file: Content/Introduction.tex
\section{Introduction}

Event sequences play a crucial role in various domains, notably in social networking~\cite{zhang2021vigdet,primack2021temporal,guo2020future} and e-commerce~\cite{gupta2023neural,zhou2023explainable}. They are instrumental in advancing recommendation systems~\cite{guo2020future,gupta2023neural} and information retrieval strategies~\cite{zhou2023explainable}. Each user action on a social network or e-commerce transaction is an event that provides insights into behavioral patterns. Analyzing these events helps predict future patterns, enhancing the predictive capabilities of recommendation systems. A key challenge in this field is accurately forecasting future occurrences based on these event patterns, which has seen significant progress recently. 

Temporal Point Processes (TPPs) \cite{daley2003introduction, aalen2008survival}, with their robust theoretical foundation, have become increasingly popular for modeling these temporal event sequences. TPPs treat time intervals between events as random variables, providing a more nuanced representation than traditional discrete-time models. The \textit{(conditional) intensity function}, a core concept in TPPs, has been parameterized in various forms, including Stationary and Non-stationary Poisson processes\cite{kingman2005p}, Hawkes processes~\cite{hawkes1971spectra, liniger2009multivariate}, and other processes~\cite{aalen2008survival,kingman1992poisson,reynaud2010adaptive}. Despite their theoretical robustness, specific TPP modeling methods sometimes struggle to align with the complexities of real-world phenomena~\cite{okawa2019deep,omi2019fully,kanazawa2023asymptotic}. This misalignment is evident in scenarios where existing models, such as the Hawkes process, which typically assumes a positive excitation effect from past events, do not accurately represent real-world dynamics, such as inhibitory effects in dietary choices~\cite{kanazawa2023asymptotic}.

To address these real-world complexities, integrating Neural Networks with Temporal Point Processes (TPPs) has been a pivotal advancement. This integration has catalyzed the development of sophisticated deep TPP models~\cite{shchur2021neural, mei2017neural,du2016recurrent}, adept at processing complex, nonlinear temporal data. Nonetheless, these advanced models confront several challenges:

\begin{itemize}[leftmargin=*]

\item \textbf{Computational Instability in Integral Computations:} Integral calculations of the intensity function pose substantial computational challenges. Efforts to accurately model this integral component for log-likelihood estimations~\cite{omi2019fully} often struggle to maintain a normalized total integral of the density function~\cite{shchur2019intensity}, underscoring the need for more stable methods.

\item \textbf{Limited High-frequency and Low-frequency Events Due to Intensity Functions Modeling:} Traditional approaches, such as relying on predetermined intensity functions that generally assume a progression post-recent events~\cite{du2016recurrent}, encounter significant challenges in high-frequency scenarios. These models often fall short on accurately capturing the complex dynamics of events, particularly when real-world data exhibits non-linear behaviors. This limitation becomes especially pronounced in scenarios where intensity changes are abrupt or involve high-frequency~(extreme)/low-frequency~(rare) events, leading to reduced predictive accuracy.

\item \textbf{Inadequacy in Handling Long-Range Temporal Dependencies:} A notable limitation in the current scope of temporal point process models is their struggle to effectively capture and predict long-range temporal dependencies. This challenge is significant in various real-world applications where understanding and forecasting extended temporal patterns are crucial, indicating a pressing need for more robust models in this domain~\cite{shchur2021neural,xiao2017wasserstein,xiao2019learning}.
\end{itemize}

In addressing these limitations, we present the \textbf{Cu}mulative Distribution \textbf{Fun}ction-based Temporal Point Process (\textbf{CuFun}), a pioneering model that innovatively focuses on directly modeling the CDF within TPPs. To our knowledge, {CuFun} is the first to employ a monotonic neural network for CDF representation, utilizing past events as a scaling factor to enhance the prediction of future events~\cite{orbach1962principles,li2008nonparametric}. This novel approach significantly improves modeling flexibility and adaptation to diverse data scenarios. \textbf{Firstly}, {CuFun} streamlines log-likelihood calculations through the use of automatic differentiation for deriving the density function, circumventing the complexities of integral computations and thereby bolstering both accuracy and numerical computational stability. \textbf{Secondly}, by focusing on CDFs through neural networks, the {CuFun} model not only transcends the constraints of traditional predefined density functions but also significantly enhances adaptability and precision in high-frequency scenarios. This approach broadens its applicability, particularly in accurately capturing the intricate dynamics of rapid event sequences. \textbf{Thirdly}, the CuFun model effectively addresses the challenge of long-range temporal dependency modeling. By leveraging the cumulative nature of its CDF-based approach, CuFun is adept at capturing intricate, extended temporal patterns, a capability that many existing TPP models lack. This proficiency enables CuFun to more accurately reflect and predict complex temporal sequences, enhancing its utility in a broad range of real-world applications where long-range periodic behaviors are pivotal.

The {CuFun} model, with its direct focus on modeling CDFs, signifies a major shift in TPPs. Its adaptability, numerical computational stability, and extensive applicability set a new benchmark in temporal event data modeling, effectively addressing the challenges inherent in traditional TPP models.
We outline our contributions:
\begin{itemize}[leftmargin=*]
\item We propose a novel TPP model based on the properties of Cumulative Distribution Functions (CDFs). This model accurately derives auxiliary functions such as intensity and density functions from the CDFs, enhancing its analytical depth.

\item We develop a novel method to incorporate past event information as scaling factors for future predictions, a feature not commonly found in traditional TPP models. This method provides deeper insight into the past events on future occurrences

\item We demonstrate the superior performance of our model through extensive experiments on both synthetic and real-world datasets, affirming its efficacy and potential as a leading method in TPP.
\end{itemize}

%% file: Content/Proposed_Model.tex
\section{Background}
Before presenting the proposed model, we first introduce preliminaries on temporal point processes.

\subsection{Temporal Point Processes Modeling}

A Temporal Point Process (TPP) is defined as a stochastic process that generates a sequence of discrete events occurring at times $\mathcal{T} = \left\{ t_1, ..., t_N \right\}$ within a specified observation window $[0, T]$. TPPs are typically characterized through a (conditional) intensity function, serving as a stochastic model to predict the timing of the next event based on previous occurrences. Specifically, within an infinitesimally small time frame [$t$, $t+dt$), and considering historical events $\mathcal{H}(t) = \left\{ t_i \in \mathcal{T} : t_i < t \right\}$, the rate of occurrence $\lambda^(t)$ for a future event, conditional on the history $\mathcal{H}(t)$ and excluding time $t$, is formally defined as follows:
\begin{equation}
\begin{aligned}
    \label{eq:intensity}
    \lambda^*(t)\mathrm{d}t
    =\lambda(t|\mathcal{H}(t))\mathrm{d}t
    =\mathbb{P}(event \in [t,t + \mathrm{d}t)|\mathcal{H}(t)).
\end{aligned}
\end{equation}
It is important to note that, consistent with prevailing literature on point processes \cite{rubin1972regular}, this paper assumes a regular point process, implying that no two events occur simultaneously. The $*$ symbol, used throughout this document, indicates a dependency on the historical context $\mathcal{H}(t)$.
Temporal Point Processes (TPPs) offer a framework for modeling event sequences, which can be equivalently represented as sequences of strictly positive inter-event times, $\tau_i = t_i - t_{i-1} \in \mathbb{R}_{+}$. The dual representations through $t_i$ (event times) and $\tau_i$ (inter-event times) are isomorphic, allowing their interchangeable application in analysis as established by \cite{shchur2019intensity}.

\subsection{Predicting Time of Next Event}
Then, we will delve into the mathematical formulations on predicting the timing of subsequent events based on historical data. 
In the TPP framework, the conditional intensity function $\lambda^{\ast}(t)$, synonymous with the hazard function $\phi^{\ast}$, plays a crucial role. It is defined in relation to the conditional density $p^{\ast}$ and its corresponding cumulative distribution function $F^{\ast}$, capturing the likelihood of an event occurring at time $t$, given the history:
\begin{equation}
\begin{aligned}
\label{eq:inten_ori}
\phi^{\ast}(\tau) = \lambda^{\ast}(t) = \frac{p^{\ast}(\tau)}{1-F^{\ast}(\tau)} = -\frac{\mathrm{d}}{\mathrm{d}\tau}\log(1-F^{\ast}\phi^{\ast}(\tau)(\tau)).
\end{aligned}
\end{equation}

Integrating this expression, and invoking the fundamental theorem of calculus, we derive the integral representation of the hazard function: 
\begin{equation}
\begin{aligned}
\label{eq:derive}
-\log(1-F^{\ast}(\tau)) = \begin{matrix}\int_{0}^{\tau} \phi^{\ast}(s)\mathrm{d}s\end{matrix},
\end{aligned}
\end{equation}
where $F^{\ast}(0)=0$ under the assumption of regularity, a standard in TPP literature. 
This leads us to an expressive formula for the CDF, $F^{\ast}(\tau)$ as:

\begin{equation}
\begin{aligned}
    \label{eq:F_right}
    F^{\ast}(\tau) = 1-\exp(\begin{matrix}-\int_{0}^{\tau}\phi^{\ast}(s)\mathrm{d}s \end{matrix}).
\end{aligned}
\end{equation}

Thus, considering Eq.~\ref{eq:intensity}, 
given the intensity function, we can obtain the conditional density function of the time $\tau_i$ until the next event by integration \cite{rasmussen2011temporal}as:
\begin{equation}
\begin{aligned}
    \label{eq:p_ori}
    {p}^*(\tau_i) 
    & = p(\tau_i|\mathcal{H}(t)) \\
    & = \lambda^*(t_{i-1} + \tau_i)\exp(-\begin{matrix} \int_{0}^{\tau_i} \lambda^*(t_{i-1} + s)\mathrm{d}s\end{matrix}).
\end{aligned}
\end{equation}

Illustrative examples of TPPs include the Stationary Poisson Process, characterized by a constant intensity function $\lambda(t|\mathcal{H}(t))=\lambda$, and the Hawkes Process, with an intensity function $\lambda(t|\mathcal{H}(t))=\mu+\sum_{i:t_i<t}g(t-t_i)$, where $\mu$ represents the base intensity and $g$ is a kernel function depicting the influence of past events. The method of intensity parametrization is widely adopted in TPPs for model learning. Here, parameters $\theta$ are estimated from an observed event sequence $\mathcal{T}$ by maximizing the log-likelihood, a pivotal concept in statistical inference for TPPs:
\begin{equation}
\begin{aligned}
    \label{eq:theta_ori}
    \theta^* 
    & = \mathrm{argmax}_\theta \begin{matrix} \sum\limits_{i} \log {p}_\theta^*(\tau_i) \end{matrix}  \\
    & = \mathrm{argmax}_\theta \left\{ \begin{matrix} \sum\limits_{i} \log \lambda_\theta^*(t_i)-{\begin{matrix} \int_{0}^{t_N} \lambda_\theta^*(s)\, \mathrm{d}s\end{matrix}} \end{matrix} \right\}.
\end{aligned}
\end{equation}

%% file: Content/methods.tex
\section{Methodology}
This section delineates the methodologies employed in our study, focusing on the intricate relationships among key functions in Temporal Point Processes (TPPs) and their implications in modeling event timings. We introduce novel approaches to parameterize the Cumulative Distribution Function (CDF) using Recurrent Neural Networks (RNN) and Monotonic Neural Networks (MNN), thereby enhancing the accuracy and numerical stability of our TPP models.

\begin{figure}[t]
\centering
\includegraphics[width=0.5\linewidth]{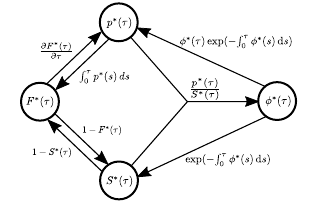}
\caption{{In Temporal Point Processes (TPPs), the Cumulative Distribution Function (CDF) $F^{(\tau)}$ is key. Derived from it are the Survivor Function $S^{(\tau)} = 1-F^{(\tau)}$, indicating the probability of no event by time $\tau$, and the Intensity Function $\lambda^{(t)} = p^{(\tau)}/S^{(\tau)}$, representing the event rate at time $t$. The Hazard Function $\phi^{(\tau)}$, equating to $\lambda^{(t)}$, reflects the immediate event occurrence risk. These relationships illustrate that deciphering the CDF enables inferring all critical TPP functions, highlighting their interconnectedness in deciphering temporal dynamics.}}
\label{fig:tpp}
\end{figure}
\subsection{Relationships among Functions in TPP}

To begin, we first explore the fundamental relationships among key functions in Temporal Point Processes (TPPs), emphasizing their interdependencies and pivotal roles in accurately modeling the timing of events.
In the realm of temporal point processes, relationships among various functions are pivotal. Given the survivor function $S^{(\tau)}$ and the density function $p^{(\tau)}$, the intensity function $\lambda^{(t)}$ is articulated as per the equation \cite{daley2003introduction}:
\begin{equation}
\begin{aligned}
\label{eq:re1}
\lambda^{(t)}= \frac{p^{(\tau)}}{S^{(\tau)}}.
\end{aligned}
\end{equation}
A key function associated with the survivor function is the cumulative distribution function (CDF). To elucidate, the interrelationship among the CDF $F^{(\tau)}$, survivor function $S^{(\tau)}$, and density function $p^{(\tau)}$ is delineated as follows:
\begin{equation}
\left\{
             \begin{array}{lr}
             S^*(\tau) = 1 - F^*(\tau), \\
             p^*(\tau) = \frac{\partial F^*(\tau)}{\partial \tau}.
             \end{array}
\right.
\end{equation}
These relationships are visually represented in Figure \ref{fig:tpp}. For in-depth derivations, refer to \cite{daley2003introduction,rasmussen2011temporal}. Figure~\ref{fig:tpp} clearly illustrates the central role of the CDF, as its derivative, when compared to the survivor function, corresponds to the intensity function.

The primary objective here is to model the CDF, accounting for historical event data. This task bifurcates into: firstly, devising a method to encapsulate past events and their integration into the main network; and secondly, ensuring the network's output corresponds to a valid CDF.

The Recurrent Neural Network (RNN) is frequently the go-to for sequential data to distill underlying patterns. In line with prevalent literature \cite{du2016recurrent,omi2019fully,shchur2019intensity}, we employ an RNN to model past event influences. Here, an input vector $x_i$ serves as the RNN's input. A straightforward representation of $x_i$ could be the inter-event interval, defined as $x_i = t_i - t_{i-1}$, or its logarithmic form $x_i = \log(t_i - t_{i-1})$. The RNN's hidden state $\boldsymbol{h_i}$ updates as follows:
\begin{align}
\label{eq:h}
\boldsymbol{h}_i = \max\left\{f(W^h\boldsymbol{h_{i-1}}+W^xx_i+b^h),0\right\},
\end{align}
where $W^h$, $W^x$, $b^h$, and $f$ represent the recurrent weight matrix, input weight matrix, bias term, and an activation function, respectively. The hidden state $\boldsymbol{h}$ of the RNN is then conceptualized as a condensed vector representation of historical events. The subsequent step involves leveraging this historical data and structuring the model to produce a valid CDF.

\subsection{Parameterizing CDF for TPP}

In this subsection, we introduce our approach to parameterizing the Cumulative Distribution Function (CDF) in TPPs, emphasizing the use of monotonic neural networks to enhance model precision. CDF modeling, devoid of integral computations and approximations, ensures numerical stability and improved effectiveness, particularly in capturing the intricate dynamics~\cite{hasan2007verification,li2008nonparametric} of rapid event sequences.

\begin{figure}[h]
\centering
\includegraphics[scale=0.9]{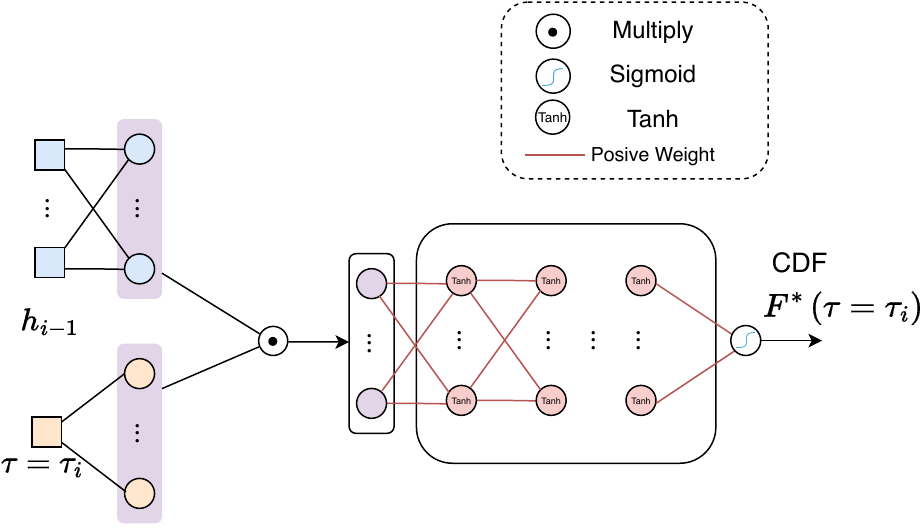}
\caption{{Details of our Monotonic Neural Network~(MNN). All connections within the MNN are constrained to be positive, 
ensuring that the derivative of the CDF with respect to $\tau$ is positive. 
The activation function for the final unit is a $sigmoid$ function, 
ensuring that the output value falls within the interval (0,1). 
These two constraints collectively guarantee that the output is a valid CDF. 
The piece-wise multiplication operation represents the scaling effect of past events on the variable $\tau$.
}}
\label{MNN}
\end{figure}

Inspired by the application of monotonic neural networks in probabilistic function modeling \cite{chilinski2020neural}, we utilize the inter-event interval as the input to a feedforward neural network. Notably, the hidden state $\boldsymbol{h_{i-1}}$ receives $x_{i-1}=t_{i-1}-t_{i-2}$ as input, which is obtained independently of the monotonic neural network. In essence, $\boldsymbol{h_{i-1}}$ represents the \textbf{covariate} variable, while $\tau_{i}$ corresponds to the \textbf{response} variable. As shown in Figure~\ref{MNN}, the cumulative distribution function is represented through a monotonic neural network ($\mathrm{MNN}(\cdot)$) as follows:
\begin{align}
\label{eq:cdf}
{F^*(\tau=\tau_{i}|\boldsymbol{h}{i-1})} = \mathrm{MNN}{\tau=\tau_{i}}(\tau,\boldsymbol{h=h_{i-1}}).
\end{align}
However, a generic neural network does not inherently ensure a valid cumulative distribution function output. To certify a function as a valid cumulative distribution function, it must adhere to constraints of monotonicity and boundedness at both positive and negative infinities, as delineated by the following three conditions:
\begin{equation}
\left\{
\begin{array}{lr}
     \textcircled{1}:  \lim\limits_{\tau \to -\infty}F(\tau| \boldsymbol{h}) = 0, \\
     \textcircled{2}: \lim\limits_{\tau \to +\infty}F(\tau| \boldsymbol{h}) = 1, \\
     \textcircled{3}: \frac{\partial F(\tau| \boldsymbol{h})}{\partial \tau} >= 0.
\end{array}
\right.
\end{equation}

\begin{figure*}[h]
\centering
\includegraphics[scale=0.9]{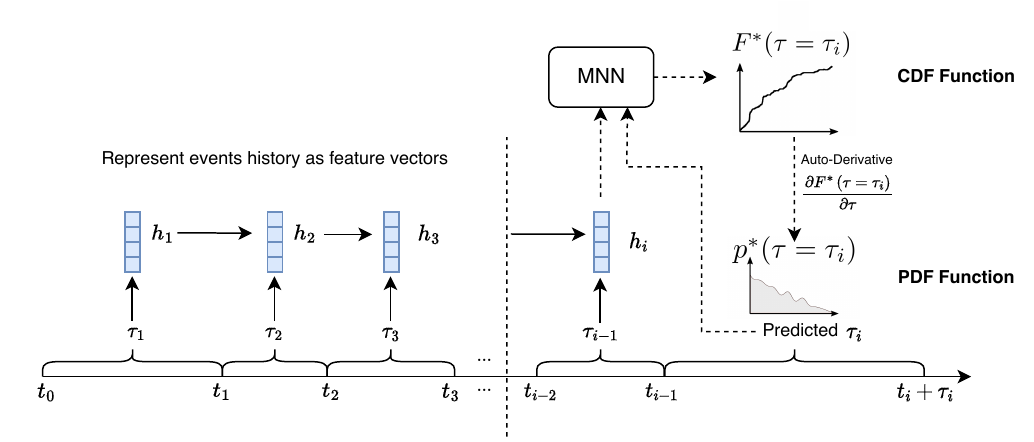}
\caption{{Model architecture. The $t_{i}$ is the time when a event happened. The time interval is denoted as $\tau_{i}$ and $\tau_{i}=t_{i}-t_{i-1}$. The density function, CDF, survivor function are denoted as $p^*$, $F^*$ and $S^*$ respectively, where $*$ symbol reminds us of dependence on past events. Past events time intervals are fed into a RNN and return the hidden vector $\boldsymbol{h}$. The future time interval $\tau$ and the hidden vector $\boldsymbol{h}$, are fed into the monotonic neural network (MNN). The output of the MNN corresponds to the CDF. Then we can further get the density function via automatic differentiation. }}
\label{whole}
\end{figure*}

We now elucidate the entire network structure and explain how these constraints are fulfilled.
The neural network (depicted in Figure \ref{whole}) consists of two primary components. Initially, as previously mentioned, past event information is encoded into a compact vector $\boldsymbol{h_{i-1}}$ via an RNN. Once $\boldsymbol{h_{i-1}}$ is obtained, it, along with $\tau_{i}$, serves as the input to the monotonic neural network.
The crux of our model lies in the second component, the monotonic neural network, highlighted in a red dotted box. Here, $\boldsymbol{h}$ and $\tau$ are processed through two distinct single-layer networks, ensuring congruent output sizes. Generally, in monotonic neural networks, these outputs are summed. However, significant magnitude disparities between the two can lead to a dominance of one value over the other. Our experiments reveal substantial differences in output magnitudes (details in the experimental section), suggesting that the influence of past events might be undervalued in future event prediction.

To address this, we replace the addition operation with an element-wise product, interpreting it as a scaling effect of historical information on future event predictions. This product then serves as the input to another feedforward neural network. In this network, the initial hidden layer is the result of the element-wise product, with each unit in the median and last layers applying activation functions $\textit{tanh}$ and $\textit{sigmoid}$ respectively, to produce the output. The distinct activation functions are denoted by varying colors in Figure \ref{whole}. Additionally, all connection weights in the monotonic neural network are constrained to positive values (negative updates during training are replaced with their absolute values), while biases remain unconstrained. Connections with positive-weight constraints are indicated by dashed lines in Figure \ref{whole}.

The compliance with the three specified constraints is examined as follows: The positive weights from $\tau$ to the final output, in conjunction with the positive attributes of $\boldsymbol{h}$ as outlined in Eq.~\ref{eq:h}, inherently satisfy constraint $\textcircled{3}$. This ensures that the output function increases monotonically with $\tau$. Moreover, employing a $sigmoid$ activation function in the final unit secures adherence to constraints $\textcircled{1}$ and $\textcircled{2}$, as inferred from $\textcircled{3}$. As a result, the network's output can approximate the cumulative distribution function. \textbf{Such a modeling approach circumvents the need for integral computations, relying instead on numerically more stable differentiation operations.}

\subsection{Loss Function}
In this subsection, we outline our strategy for optimizing the model and designing the loss function, with a focus on deriving the density function from the Cumulative Distribution Function (CDF). This step is vital for precise event prediction within Temporal Point Processes (TPPs). By obtaining the CDF, we can then seamlessly compute the density function through automatic differentiation with respect to $\tau$. This process is facilitated by frameworks such as PyTorch and TensorFlow, as demonstrated below:
\begin{equation}
    \label{eq:p_cufun}
     {p}^*(\tau=\tau_i) = \frac{\partial {F^*(\tau=\tau_{i}|\boldsymbol{h}}_{i-1})}{\partial \tau}.
\end{equation}
Compared to (\ref{eq:p_ori}), density function of this formation is significantly conciser and easier to estimate.
Now we can further use the $F^*(\tau_{i})$ and $p^*(\tau_{i})$ to derive other functions such as intensity function through relationships shown in Figure \ref{fig:tpp}:

\begin{align}
    \label{eq:lambda_cufun}
    {\lambda(t_{i})} = \frac{(1-F(\tau_{i}|\boldsymbol{h}_{i-1}))}{p(\tau_{i}|\boldsymbol{h}_{i-1})}.
\end{align}
The log-likelihood function can be obtained as:
\begin{equation}
\begin{aligned}
    \label{eq:nll}
    \log{\left\{ {p}_\theta^*(\tau_i) \right\} } 
    = \begin{matrix} \sum\limits_{i} \log {p}_\theta^*(\tau_i) \end{matrix} 
    = \begin{matrix} \sum\limits_{i} \log \frac{\partial {F(\tau=\tau_{i}|\boldsymbol{h}}_{i-1})}{\partial \tau}.
    \end{matrix}
\end{aligned}
\end{equation}
Compared to (\ref{eq:theta_ori}), this log-likelihood term dispense with the intractable integral part and can be exactly evaluated even for a complex model via a simple monotonic neural network. Through the monotonic neural network for estimating cumulative distribution function $F_{\theta}^*(\tau|\boldsymbol{h})$ and
given a sequence of observations $\mathcal{T}$ , the parameters $\theta$ can be estimated by maximizing the log-likelihood:
\begin{equation}
\begin{aligned}
    \label{eq:theta}
    \theta^* 
    & = \mathrm{argmax}_\theta \left\{ \begin{matrix} \sum\limits_{i} \log \frac{\partial {F(\tau=\tau_{i}|\boldsymbol{h}}_{i-1})}{\partial \tau}
    \end{matrix} \right\}.
\end{aligned}
\end{equation}

%% file: Content/Experiment.tex
\section{Experiment}

In this section, we conduct the log-likelihood comparison experiment on both synthetic and real-world datasets.  In addition, for better demonstrating the effectiveness and superiority of our model, we also compare the intensity function predictive performance. In order to show the advance of element-wise product, we visualize and compare the corresponding values when using element-wise product and addition during training. 

\subsection{Dataset Description}
We examine 2 synthetic datasets and 8 real-world datasets from various domains.

\subsubsection{Synthetic Datasets}
The synthetic datasets are created as: 
\quad\\
\noindent \textbf{Hawkes Processes:}
The Hawkes process employs a conditional intensity function $\lambda(t|H(t))=\mu+{\begin{matrix} \sum\limits_{j=1}^M \alpha_j \beta_j \exp\left\{ -\beta_j(t-t_j) \right\} \end{matrix}}$. In the \textit{Hawkes1} dataset, the parameters are set as $M=1$, $\mu=0.2$, $\alpha_1=0.8$, and $\beta_1=0.8$. The \textit{Hawkes2} dataset features $M=2$, $\mu=0.2$, $\alpha_1=0.4$, $\alpha_2=0.4$, $\beta_1=1.0$, and $\beta_2=20.0$.

\noindent \textbf{Renewal Processes:}
The intervals $\left\{ \tau_i=t_{i+1}-t_{i} \right\}$ in stationary renewal processes are independent and identically distributed according to $p(\tau)$, a probability density function. We utilize a log-normal distribution with a mean of $1.0$ and a standard deviation of $6.0$ for $p(\tau)$, following \cite{omi2019fully}. For non-stationary renewal processes 
\cite{lindqvist2003trend}, the sequence $\left\{t_i \right\}$ is derived by initially generating a sequence $\left\{ {t_i^\prime} \right\}$ from a stationary process and then rescaling it with ${t_i^\prime} = \int_{0}^{t_i}r(t), dt$, where $r(t)$ is a non-negative trend function. The stationary process uses a gamma distribution with a mean of $1.0$ and a standard deviation of $0.5$, and $r(t)=0.99sin(2\pi t/20000)+1$ for the trend function.

\subsubsection{ Real-world Datasets}
\quad\\
\noindent \textbf{Stack Overflow:}
Stack Overflow, a renowned Q\&A website for programming and development, utilizes a badge system to foster user engagements~\cite{du2016recurrent,grant2013encouraging}. The dataset focuses on the most active users, tracking their earned repeatable badges, which reflect ongoing community contributions.

\noindent \textbf{MIMIC2\footnote{https://github.com/musically-ut/tf\_rmtpp\label{web2}}:}
The MIMIC2 medical dataset comprises anonymized records of patient visits to the Intensive Care Unit, where each event is marked to denote a hospital visit.

\noindent \textbf{Twitter\textsuperscript{\ref{web2}}:}
The Twitter dataset includes sequences of retweets, each associated with an original tweet and time-stamped relative to the original tweet's creation.

\noindent \textbf{Reddit\footnote{https://github.com/srijankr/jodie/\label{web1}}:}
The Reddit dataset features user submissions on various subreddits~\cite{kumar2019predicting,shchur2019intensity}, focusing on active subreddits and posts from prominent users. This dataset is particularly notable for its focus on active subreddits, which are likely to have a er degree of user engagement and interaction.

\noindent \textbf{Wikipedia\textsuperscript{\ref{web1}}:}
This dataset includes the most edited Wikipedia pages, documenting edits by ly active users.

\noindent \textbf{Music:}
The Music dataset \cite{celma2010music} tracks user-specific music listening sequences, with each event detailing the listening time.

\noindent \textbf{Bookorder\textsuperscript{\ref{web1}}:}
Bookorder contains high-frequency trading data from the New York Stock Exchange, including transaction details such as time (in milliseconds) and action (buy or sell).

\noindent \textbf{Mooc\textsuperscript{\ref{web1}}:}
This dataset comprehensively records student interactions with an online course system. It includes detailed activities such as video viewing and question-solving, which tracks student engagement and response patterns.

\noindent \textbf{Yelp\footnote{https://www.yelp.com/dataset/challenge}:}
Yelp dataset comprises a vast array of reviews and business information from Yelp's review platform. Specifically, we focus on the reviews of the top 300 most popular restaurants in Toronto, compiling a sequential dataset of feedback for each restaurant to observe the temporal evolution of customer trends over time.

\begin{figure*}[!htp]
\centering
\begin{subfigure}{.225\textwidth}
  \centering
  \includegraphics[width=\linewidth]{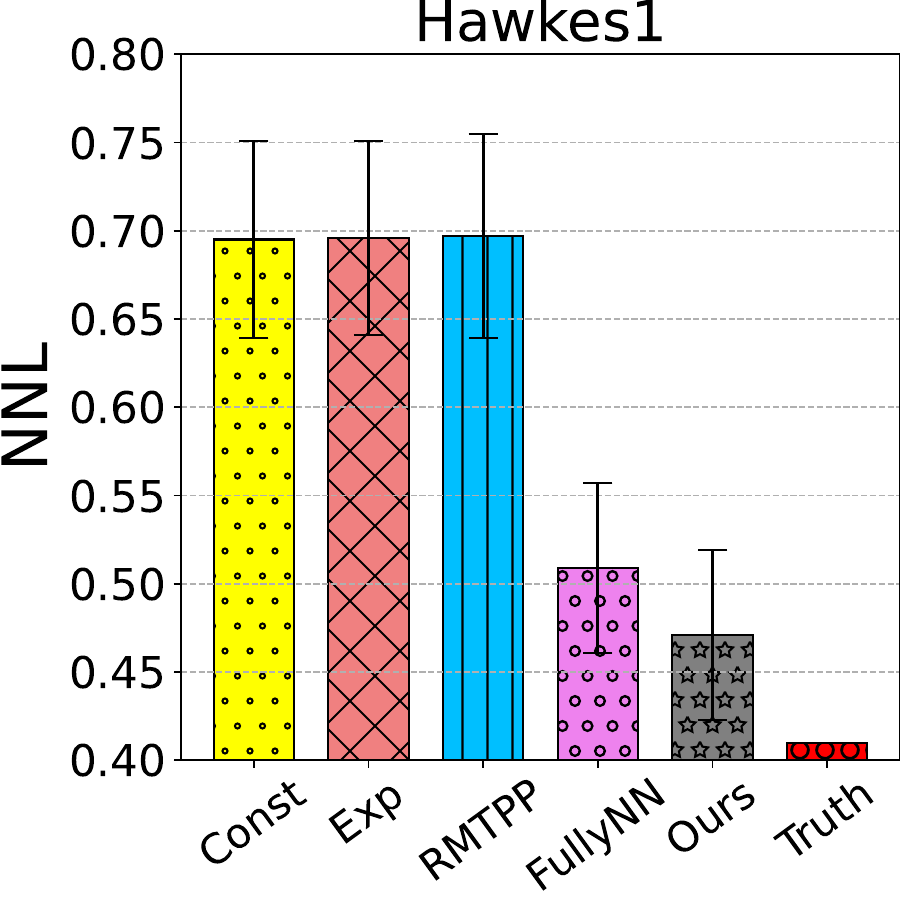}
\end{subfigure}%
\begin{subfigure}{.225\textwidth}
  \centering
  \includegraphics[width=\linewidth]{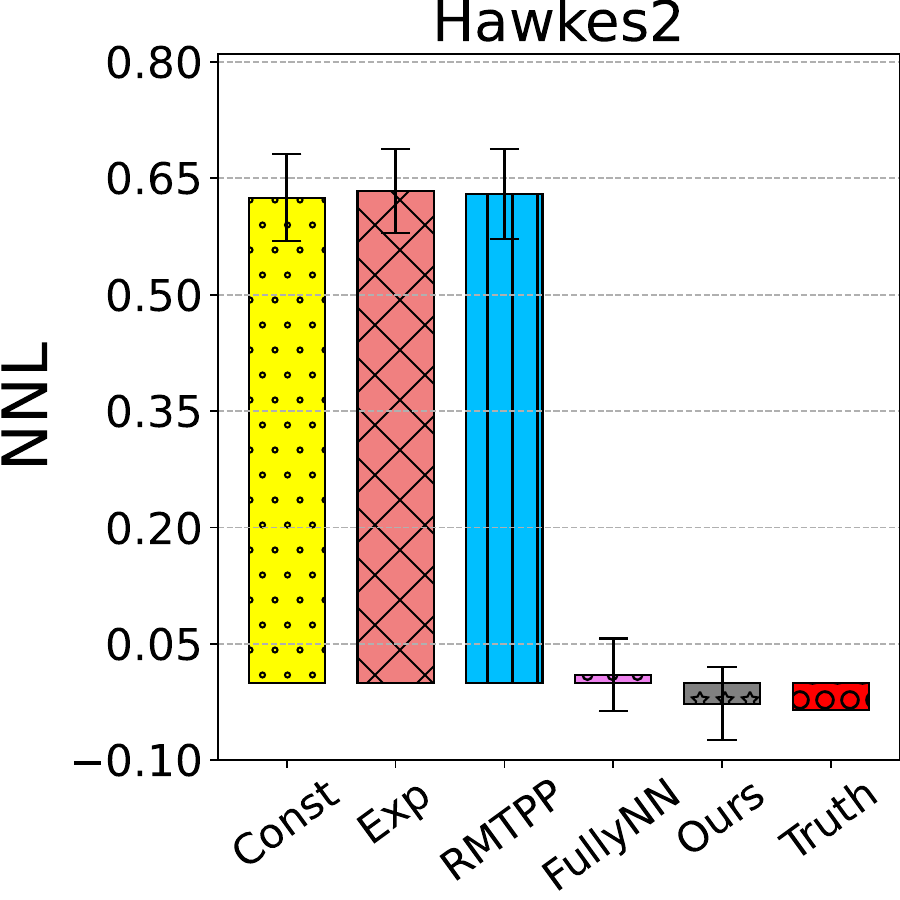}
\end{subfigure}%
\begin{subfigure}{.225\textwidth}
  \centering
  \includegraphics[width=\linewidth]{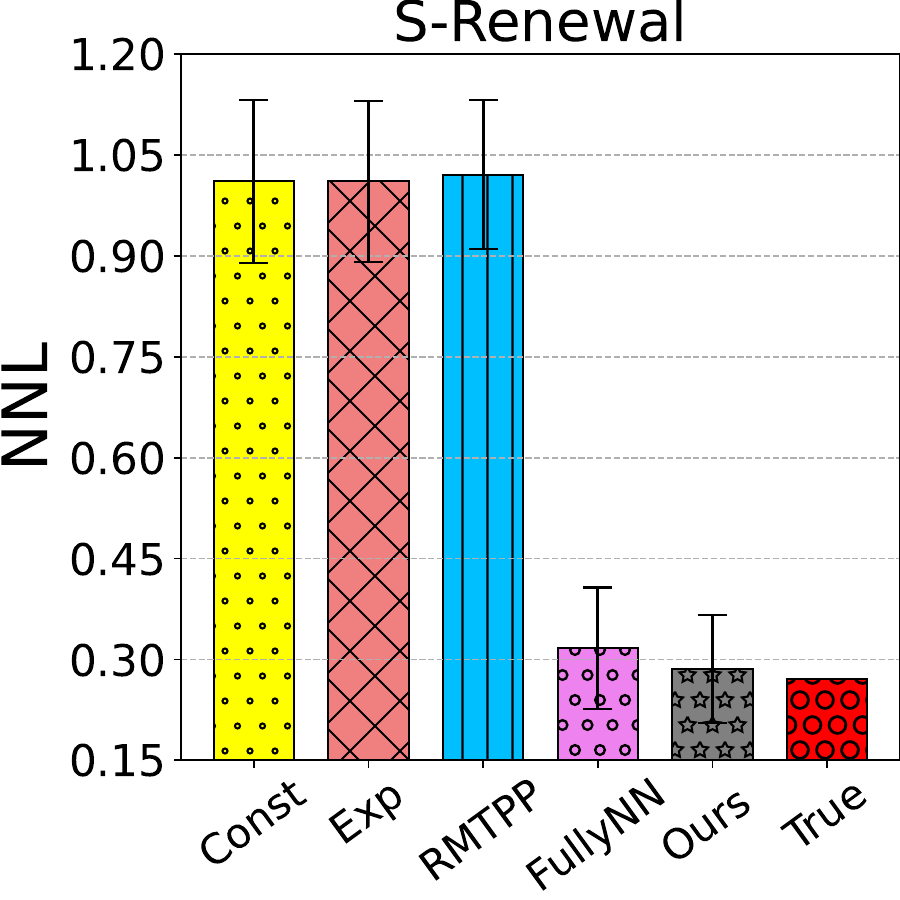}
\end{subfigure}%
\begin{subfigure}{.225\textwidth}
  \centering
  \includegraphics[width=\linewidth]{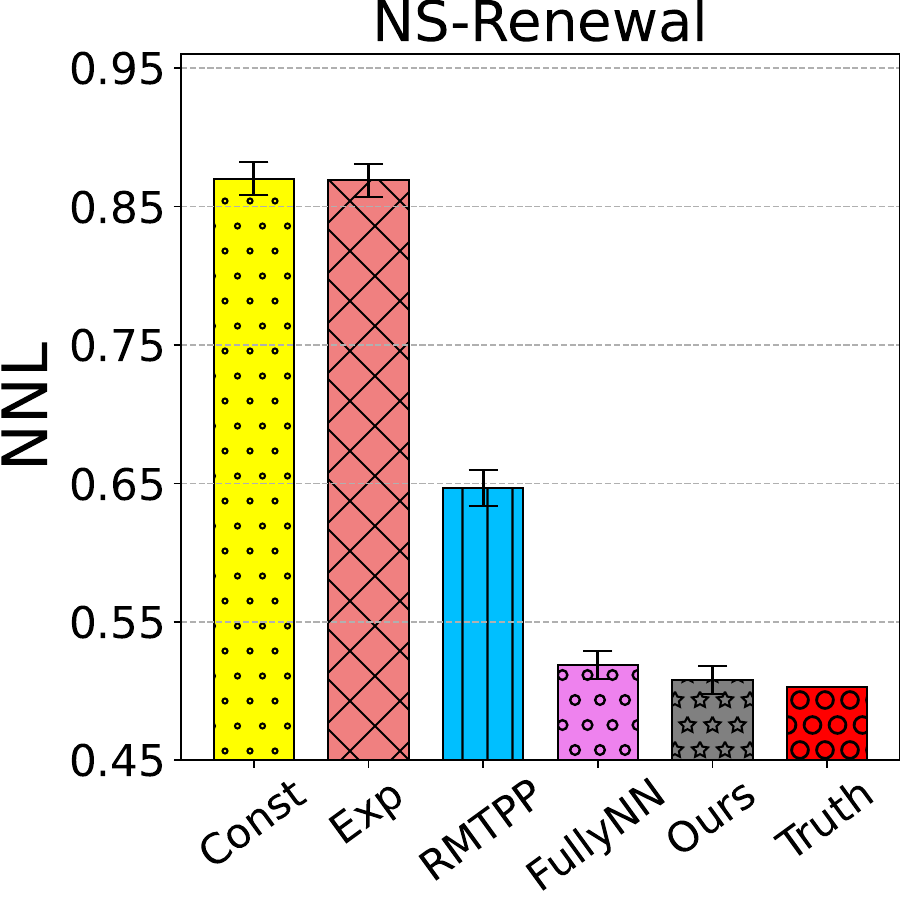}
\end{subfigure}%
\newline
\caption{NLL comparison on synthetic datasets. \textbf{Lower score is better}. All the improvements of our CuFun are \textbf{statistically significant} (i.e., two-sided t-test with $p$ < 0.05) over baselines).}
\label{nll_synthetic}
\end{figure*}

\subsection{Baselines}
In our experiment, we evaluate several baseline models:
\begin{itemize}[leftmargin=*]
    \item \textbf{Constant (Const) Model}\cite{sill1997monotonic,omi2019fully}: This model, built on a Recurrent Neural Network (RNN), assumes a constant intensity function over time. It simplifies the modeling process by maintaining a steady exponential distribution for inter-event intervals, whose mean is determined by the RNN's hidden state.
    
    \item \textbf{Exponential (Exp) Model}: In contrast to the Constant Model, this model uses an RNN to develop an intensity function that varies exponentially with time interval $\tau$. This allows for a dynamic representation of the intensity, shaped by parameters.
    
    \item \textbf{RMTPP Model}\cite{du2016recurrent}: The RMTPP employs a recurrent neural network to capture the dynamics of temporal point processes. It uniquely models the intensity function as a non-linear representation of the event history, facilitating effective modeling without specific parametric constraints.
    
    \item \textbf{FullyNN Model}\cite{omi2019fully}: This model leverages a feedforward neural network to model the integral intensity function. It excels in flexible modeling and precise log-likelihood evaluation without numerical approximations but does \textbf{NOT} incorporate modeling of the Cumulative Distribution Function (CDF).
\end{itemize}

\subsection{Implementation Details}

In our experimental setup, each test is repeated ten times with different splits of the train/validation/test sequences, ensuring that the length of all sequences is constrained to a maximum of 128. The model parameters are optimized by minimizing the negative log-likelihood (NLL) of the training sequences. During this process, we compute and record the mean NLL value, along with its standard deviation, for the test sequences. For optimization, we utilize the Adam algorithm \cite{kingma2014adam} with a learning rate of \(10^{-3}\). Each mini-batch comprises 64 sequences. The training is conducted for up to 3000 epochs, and the optimization is halted if there is no improvement observed for a continuous span of 100 epochs. To maintain fairness in comparison, all hyperparameters, including the \(L_2\) regularization value, are uniformly set across all tests. Specifically, for the LogNormMix model, we choose the number \(K\) of mixture components to be 64, as this configuration has been identified to yield the best performance in previous studies \cite{shchur2019intensity}. Our experiments are meticulously conducted on the PyTorch~\footnote{https://pytorch.org/}, leveraging its robust computational capabilities and flexible framework.
Moreover, we utilize the Negative Log-Likelihood (NLL) as a metric to evaluate the performance of our probabilistic models. NLL quantifies the model's proficiency in predicting observed data, penalizing deviations from actual values. Formally, for a probabilistic model \( P \) assigning probabilities to the observed data \( \{y_i\}_{i=1}^n \) based on parameters \( \theta \), the NLL is defined as:
\begin{equation}
NLL(\theta) = -\sum_{i=1}^n \log P(y_i | \theta)
\end{equation}
Here, \( \log \) represents the natural logarithm. In our analysis, a lower NLL value indicates a better fit of the model to the data, making it a valuable metric for comparing models.

\begin{figure*}[!tp]
\centering
\begin{subfigure}{.225\textwidth}
  \centering
  \includegraphics[width=\linewidth]{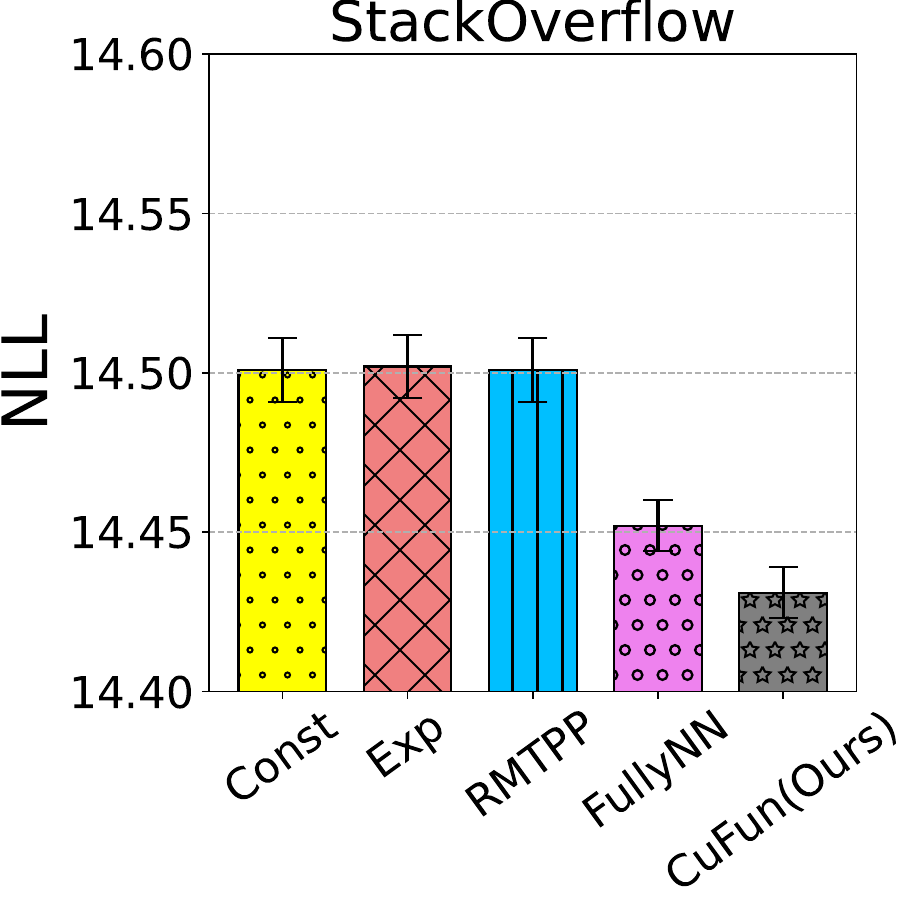}
\end{subfigure}%
\begin{subfigure}{.225\textwidth}
  \centering
  \includegraphics[width=\linewidth]{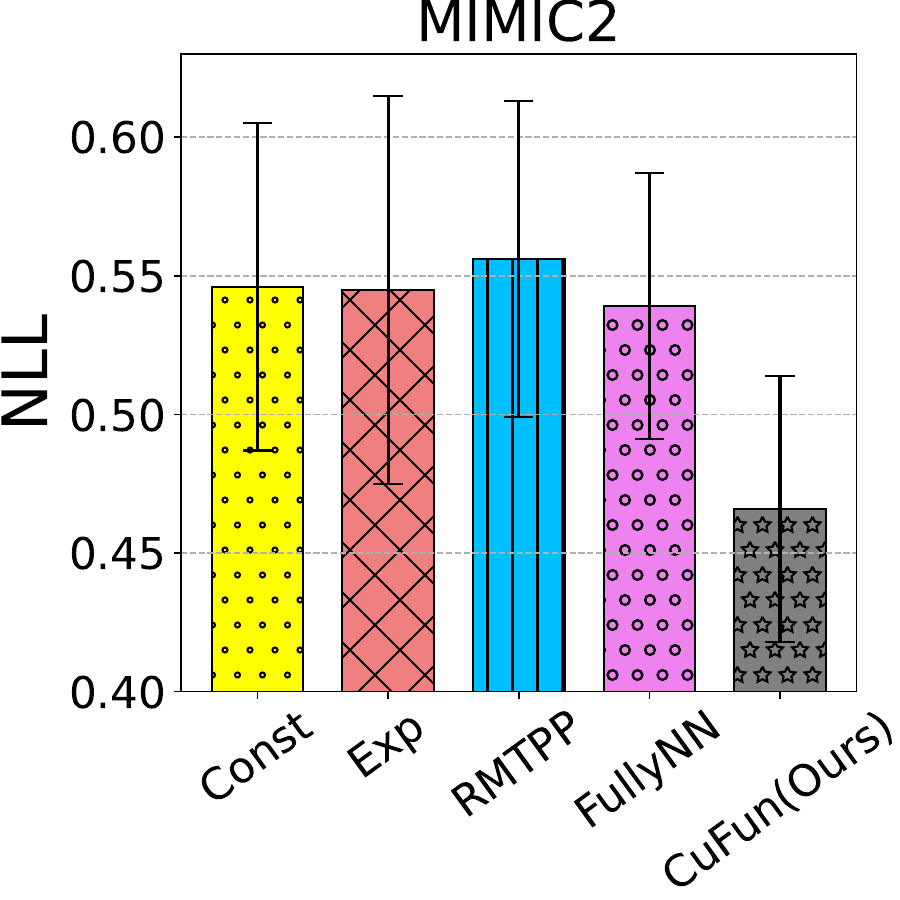}
\end{subfigure}%
\begin{subfigure}{.225\textwidth}
  \centering
  \includegraphics[width=\linewidth]{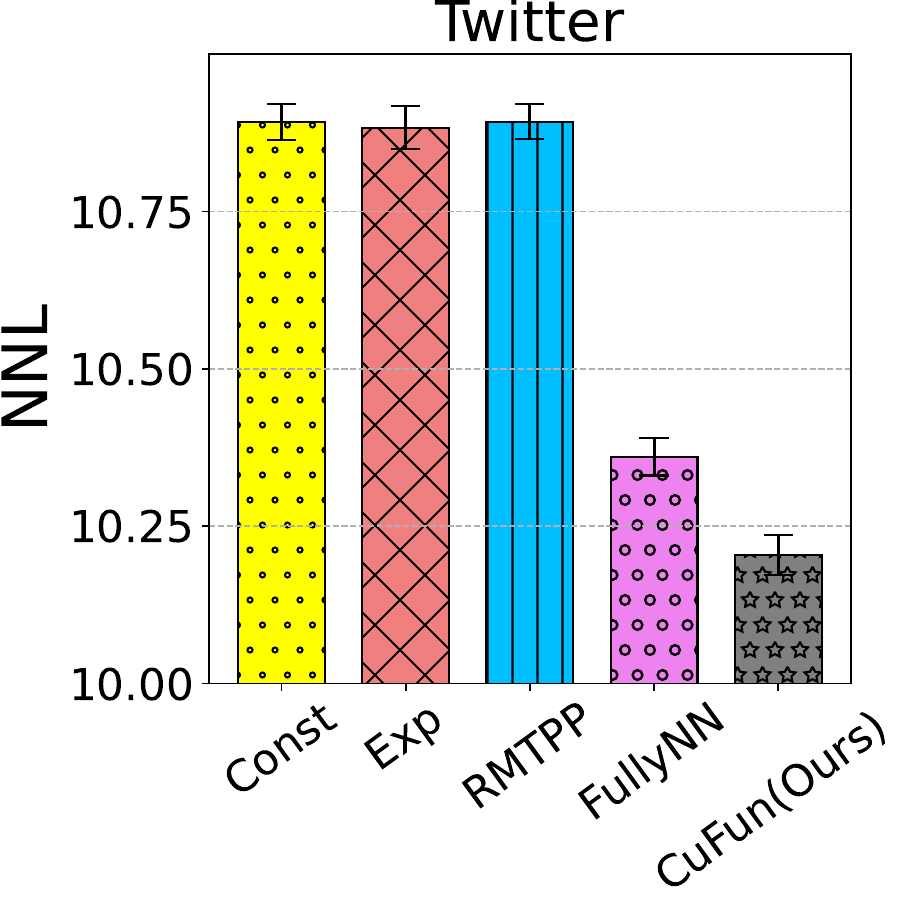}
\end{subfigure}%
\begin{subfigure}{.225\textwidth}
  \centering
  \includegraphics[width=\linewidth]{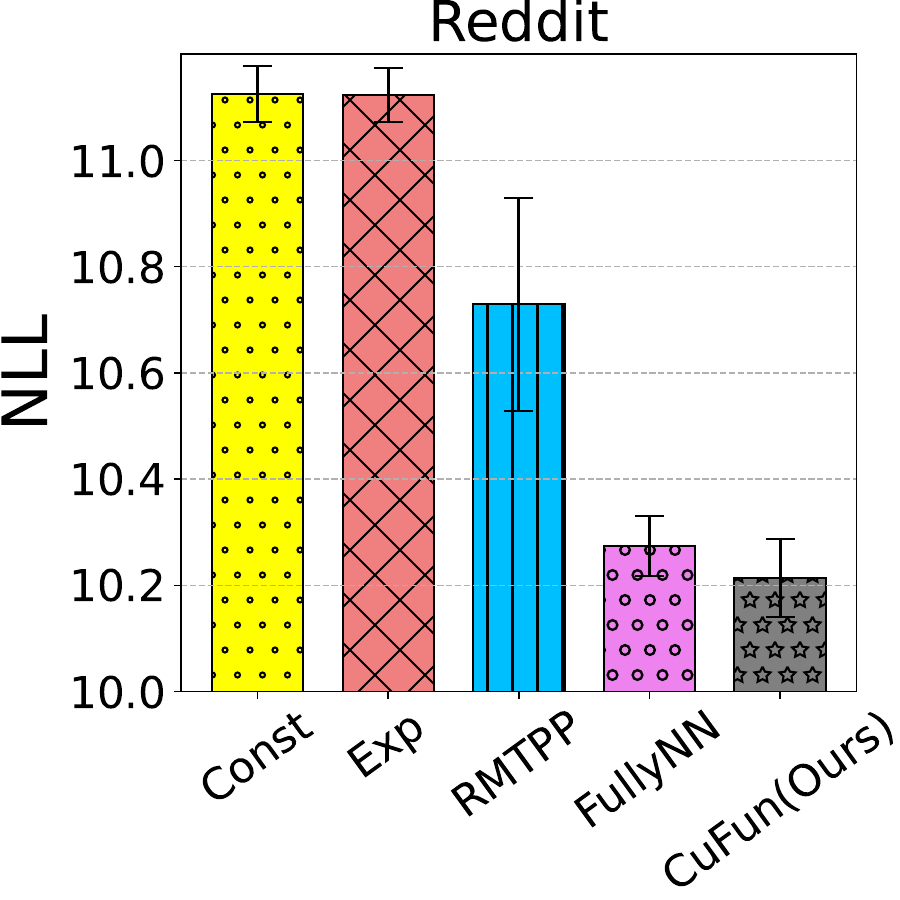}
\end{subfigure}%
\newline
\begin{subfigure}{.225\textwidth}
  \centering
  \includegraphics[width=\linewidth]{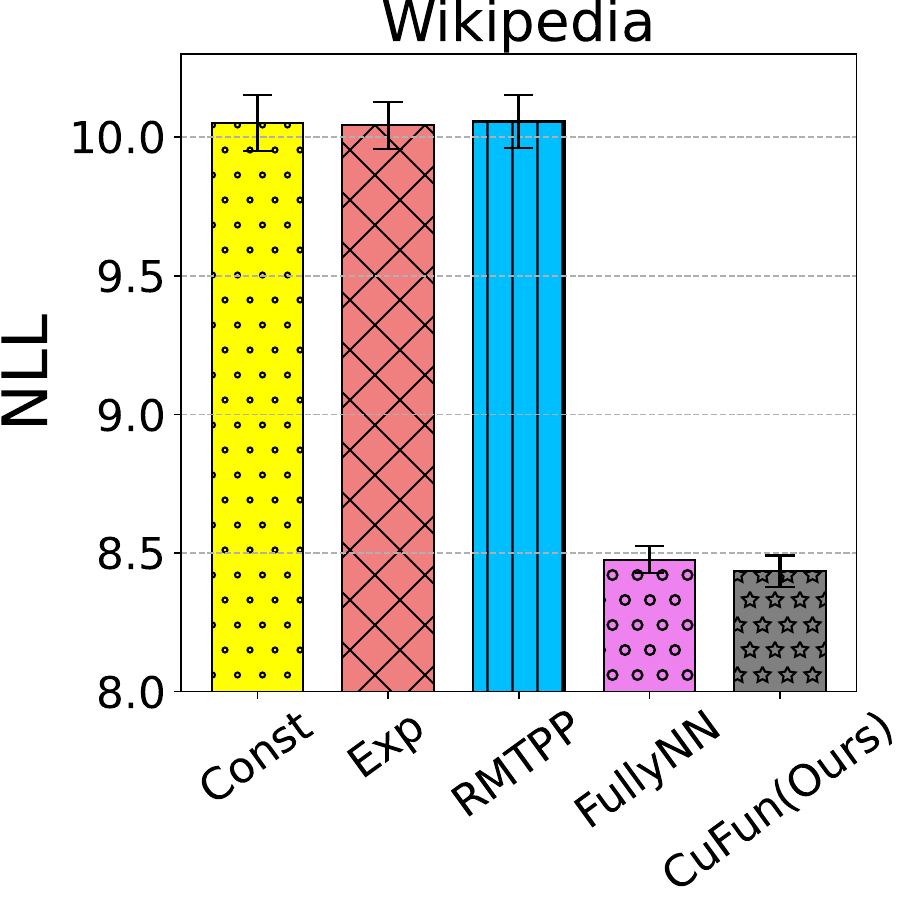}
\end{subfigure}%
\begin{subfigure}{.225\textwidth}
  \centering
  \includegraphics[width=\linewidth]{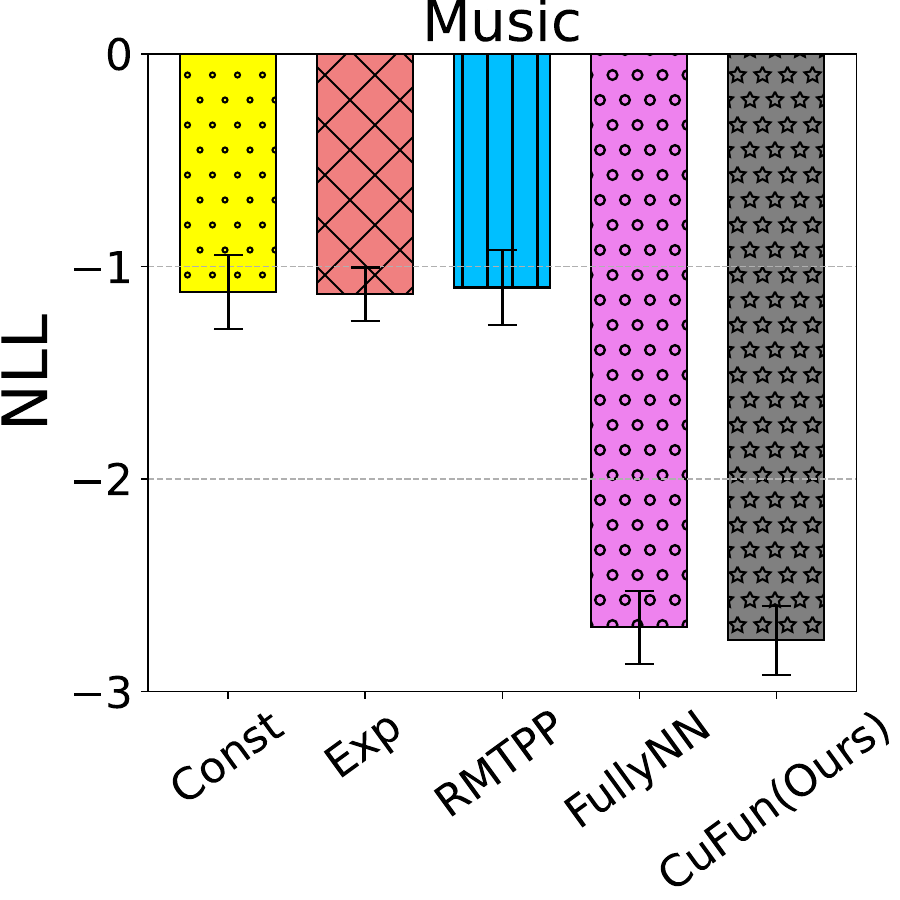}
\end{subfigure}%
\begin{subfigure}{.225\textwidth}
  \centering
  \includegraphics[width=\linewidth]{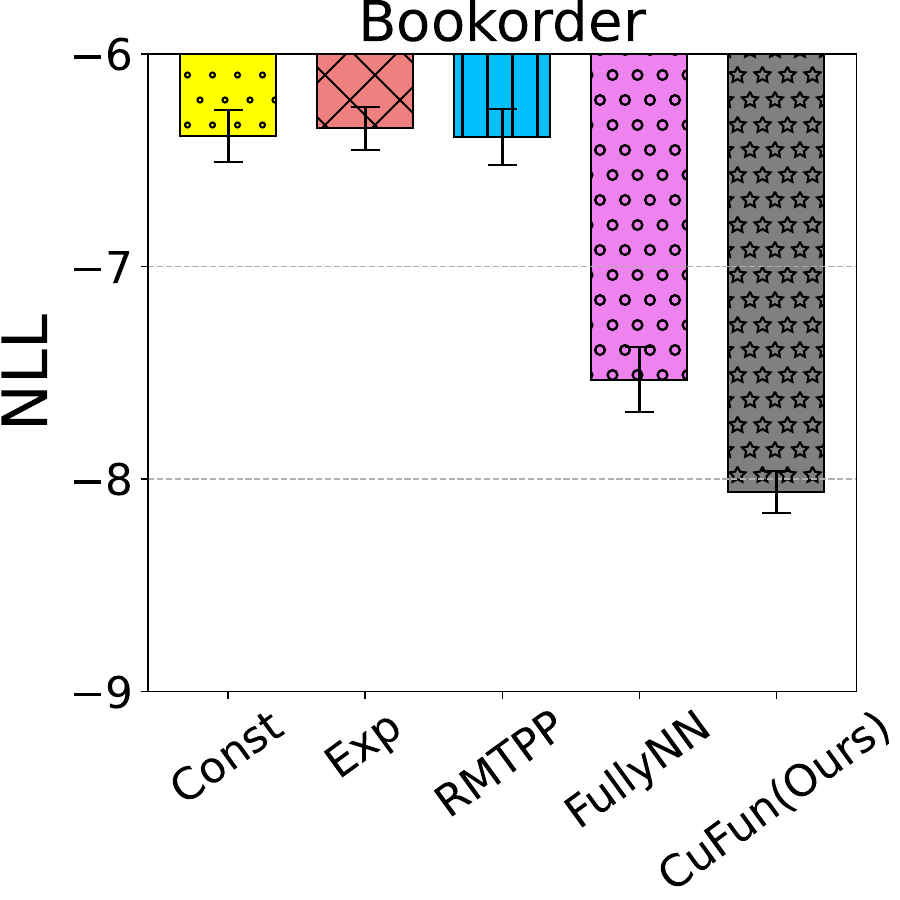}
\end{subfigure}%
\begin{subfigure}{.225\textwidth}
  \centering
  \includegraphics[width=\linewidth]{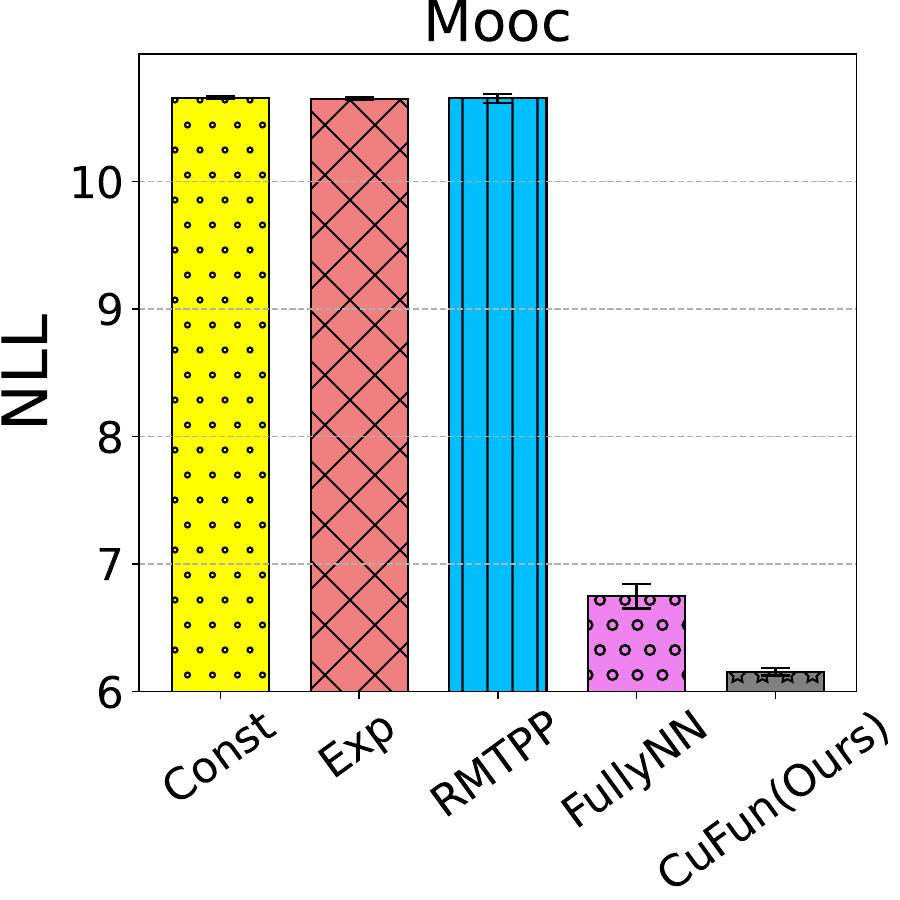}
\end{subfigure}%
\newline
\caption{NLL comparison on real-world datasets. Each score is obtained by subtracting the absolute score of our model.  \textbf{Lower score is better}. These results collectively illustrate the robustness and versatility of the CuFun model affirming its advantages in handling various types of temporal point process data. All the improvements(except Wikipedia and Music) of our CuFun are \textbf{statistically significant} (i.e., two-sided t-test with $p$ < 0.05) over baselines).}
\label{nll_real}
\end{figure*}

\subsection{NLL Comparison on Synthetic Datasets}

In this experiment, our analysis initially focuses on synthetic datasets, which serve as a controlled environment to rigorously test and validate the performance of our models, setting a foundational comparison before progressing to real-world data in the subsequent section. However, are shown in Figure \ref{nll_synthetic}, it is important to note these findings as preliminary, paving the way for further insights from subsequent real-world data analysis:

Firstly, we observed that the CuFun model consistently demonstrated superior performance across all datasets. Specifically, the NLL values for the CuFun model on the Hawkes1, Hawkes2, S-Renewal, and NS-Renewal datasets were 0.471, -0.027, 0.286, and 0.508, respectively, all of which were lower than those of other baseline models, particularly when compared to the FullyNN model (with NLL values of 0.509, 0.010, 0.317, and 0.519). This result indicates that CuFun's approach of directly modeling the CDF effectively captures the complex dynamics of temporal point processes, leading to more accurate predictions.

Secondly, we noted that although the FullyNN model demonstrated significant prowess in differential computation, especially on the Hawkes1 and S-Renewal datasets, CuFun outperformed it by integrating differential computation with CDF modeling. On the Hawkes1 and S-Renewal datasets, the NLL values for CuFun were 0.471 and 0.286, respectively, lower than those of the FullyNN model, further proving CuFun's superiority in adapting to different data scenarios and enhancing predictions.

Finally, when comparing the CuFun model with traditional TPP models, CuFun exhibited significant advantages in modeling flexibility and applicability. For instance, compared to the RMTPP model, CuFun's NLL values were closer to the true model-generated data (with NLL values of 0.410, -0.035, 0.271, and 0.503) on the Hawkes1, Hawkes2, S-Renewal, and NS-Renewal datasets. This result underscores the superiority of CuFun in directly modeling the CDF, which not only simplifies the computation process but also broadens the model's applicability, making it a more versatile and powerful tool for temporal event data modeling.

In summary, through a detailed analysis of the CuFun model's performance on synthetic datasets, we can draw a clear conclusion: the CuFun model shows significant advantages in handling the complexities of temporal point processes. Its unique approach to CDF-based modeling not only improves the accuracy of predictions but also enhances the flexibility and applicability of the model.

\subsection{Comparison on Real-world Datasets}

As shown in Figure~\ref{nll_real}, in the evaluation of real-world datasets from varied domains such as Twitter, Wikipedia, Reddit, Mooc, Music, MIMIC2, Bookorder, and StackOverflow, the CuFun model exhibited consistent superiority over other models in terms of negative log-likelihood (NLL) values. Specifically notable were the datasets like Bookorder (NLL: -8.060) and Mooc (NLL: 6.155), where CuFun showed significant advantages, indicating its  flexibility and accuracy in handling diverse temporal event data. Directly modeling the Cumulative Distribution Function (CDF) in CuFun may contribute to more stable numerical behavior, particularly for \textbf{extreme or rare events}, making it ly suitable for TPPs~\cite{magdon1998neural}.

Furthermore, in datasets characterized by high-frequency events, such as Bookorder and Mooc, CuFun (Bookorder NLL: -8.060, Mooc NLL: 6.155) markedly outperformed the FullyNN model (Bookorder NLL: -7.532, Mooc NLL: 6.749) and other competitors. This underscores CuFun's enhanced capability in accurately capturing and \textbf{predicting extreme~(high-frequency) events.} The adoption of CDF modeling in CuFun likely provides a more stable and precise numerical performance.

In the context of social media and Q\&A website data, such as Twitter (NLL: 10.2043) and Stack Overflow (NLL: 14.431), all models performed similarly, underscoring the adaptability of CuFun in these interactive event settings. The results emphasize the applicability of CuFun in accurately modeling social interactions and activities, with its CDF modeling strategy potentially offering superior predictive performance.
Additionally, CuFun's exceptional performance was also evident in the medical and music datasets, MIMIC2 (NLL: 0.466) and Music (NLL: -2.759), respectively. These results light CuFun's effectiveness in diverse domains, including medical data analysis and personal music listening behavior modeling. The direct modeling approach of CDF within CuFun may lead to more stable and accurate predictions in these fields.

These findings collectively illustrate the robustness and versatility of the CuFun model across a spectrum of real-world datasets, affirming its advantages in handling various types of temporal point process data, with a special emphasis on its stability and accuracy, especially in scenarios involving extreme or rare events. This also showcases the potential role of CDF modeling in enhancing recommendation systems and information retrieval in event sequence.

\begin{figure}[t]
\centering
\includegraphics[width=0.7\linewidth]{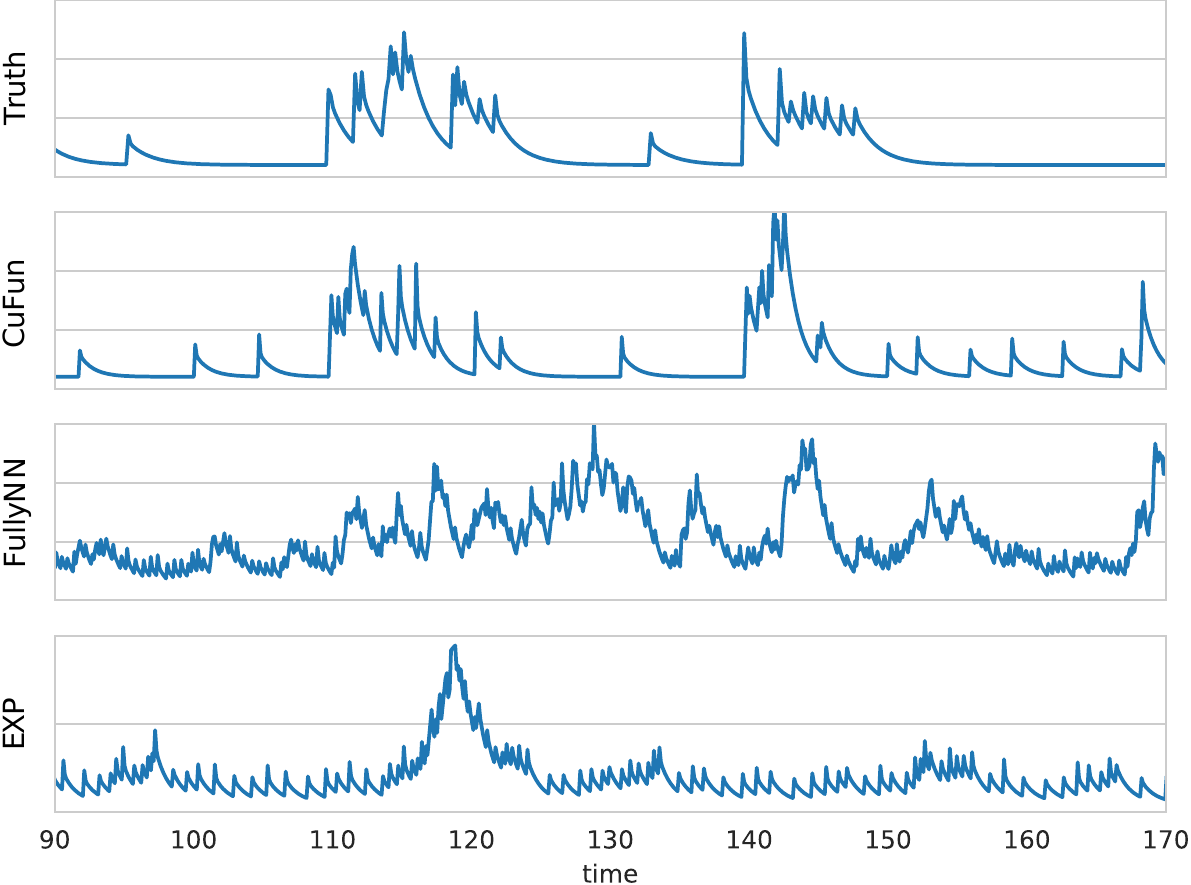}
\caption{{The intensity functions estimated from different models. 
This reveals the instability in EXP's integral calculations and the inadequacy of the Fully model in high-frequency scenarios, indicating our superiority.}}
\label{simu_haw2}
\end{figure}
\subsection{Intensity Function Estimation}
\subsubsection{Discussion: High-Frequency Modeling on Hawkes1 Dataset}

The intensity function holds a pivotal role in Temporal Point Processes (TPPs), with the Hawkes process~\cite{hawkes1971spectra} being fundamental in modeling the mutually and self-exciting behavior of events. Accurately modeling the intensity function of the Hawkes Process is crucial for practical applications. In this task, we concentrated on the intensity function estimation and compared our model, CuFun, with the FullyNN model and the Exp model as a baseline, using the Hawkes2 dataset. Other synthetic datasets were excluded due to the lack of obvious model differences in certain processes, such as the self-correcting process~\cite{isham1979self}~\cite{omi2019fully}. Additionally, we included the true model, i.e., the model generating the data, as a reference point.

Figure \ref{simu_haw2} illustrates the intensity results from different models. The true model is depicted by a grey solid line, while the other models are represented by dashed lines in various colors. The orange, blue, and green dashed lines represent the CuFun, FullyNN, and Exp models, respectively.  
In our analysis, it was observed that the Exp model falls short in accurately modeling Hawkes process events, \textbf{primarily due to the approximation of integrals}. Conversely, both the CuFun and FullyNN models exhibit a significantly better fit for appropriate intensity changes. Additionally, while the Fully model does not employ approximation of integrals, it is marked by frequent spikes and a lack of smoothness, notably in high-frequency scenarios, thereby failing to precisely reflect the original distribution's characteristics.
This observation not only lights the superior performance of the CuFun model but also indicates that direct modeling of the cumulative hazard function may not be suitable for high-frequency event modeling. Directly modeling the CDF, as employed in CuFun, might lead to more stable numerical behavior, \textbf{particularly in scenarios involving extreme or rare events.} Therefore, this approach is particularly advantageous for Temporal Point Processes (TPPs), as it adeptly handles the intricacies of time-based event data. It also proves beneficial in recommendation and information retrieval scenarios, where its capacity for detailed temporal analysis can enhance accuracy and relevance~\cite{magdon1998neural}.

\begin{figure}[t]
\centering
\includegraphics[width=0.7\linewidth]{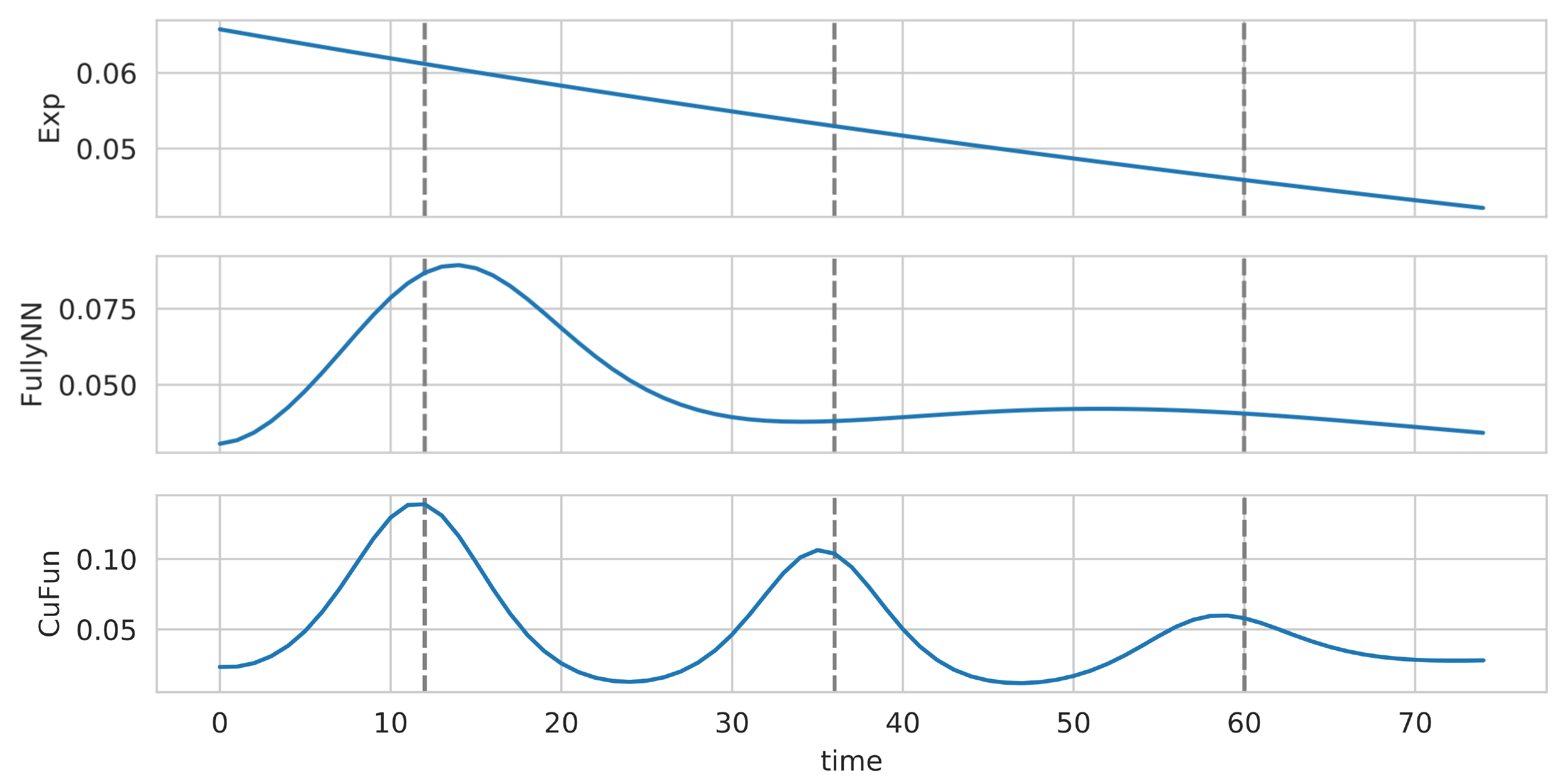}
\caption{{Comparative Analysis of Model Performance in Density Prediction on Yelp Dataset. This graph illustrates the density curves for each model, specifically emphasizing their long-range modeling capabilities in handling temporal event sequences on Yelp Dataset.}}
\label{density_yelp}
\end{figure}
\subsubsection{Discussion: Long-Range Modeling on Yelp Dataset}
We investigated various models' ability to capture long-range dependencies in event sequences using the Yelp dataset, characterized by periodic check-in events. This dataset serves as a benchmark to assess models' performance, especially in replicating long-range, periodic patterns. The NLL values for the Yelp dataset show CuFun's superiority with a score of 12.70, compared to FullyNN (13.04), RMTPP (13.36), Exp (13.35), and Const (13.36). These results highlight CuFun's strong performance.
The density prediction for the Yelp dataset, depicted in Figure ~\ref{density_yelp}, reveals significant performance differences among the models. Notably, the Exponential (Exp) model shows a marked inability to capture the dataset's periodic nature

This shortfall lights the limitations of some traditional models in dealing with complex, long-range temporal dependencies.
In contrast to the Exp model, both the Fully Neural Network (FullyNN) and our CuFun model exhibit the ability to replicate the periodic pattern of the dataset. However, it is particularly noteworthy that the CuFun model outperforms the FullyNN model in this aspect. This superior performance of CuFun underscores its efficacy in modeling the Cumulative Distribution Function (CDF) and its competence in evaluating density functions. The success of CuFun in capturing the inherent periodicity of the event sequences in the Yelp dataset suggests that its CDF-based approach is more adept at modeling long-range dependencies compared to the hazard function approach employed by FullyNN.
 \textbf{The implications of these findings are significant for temporal modeling, especially in contexts where long-range periodic behavior dependencies are prevalent. The CuFun model's ability to more accurately represent and predict such patterns demonstrates its potential as a powerful tool in temporal point process analysis.} This capability is especially relevant in real-world cases like social media analysis, e-commerce trends, and domains where forecasting periodic or cyclic behaviors is crucial.

\begin{figure}[t]
\centering
\includegraphics[width=0.7\linewidth]{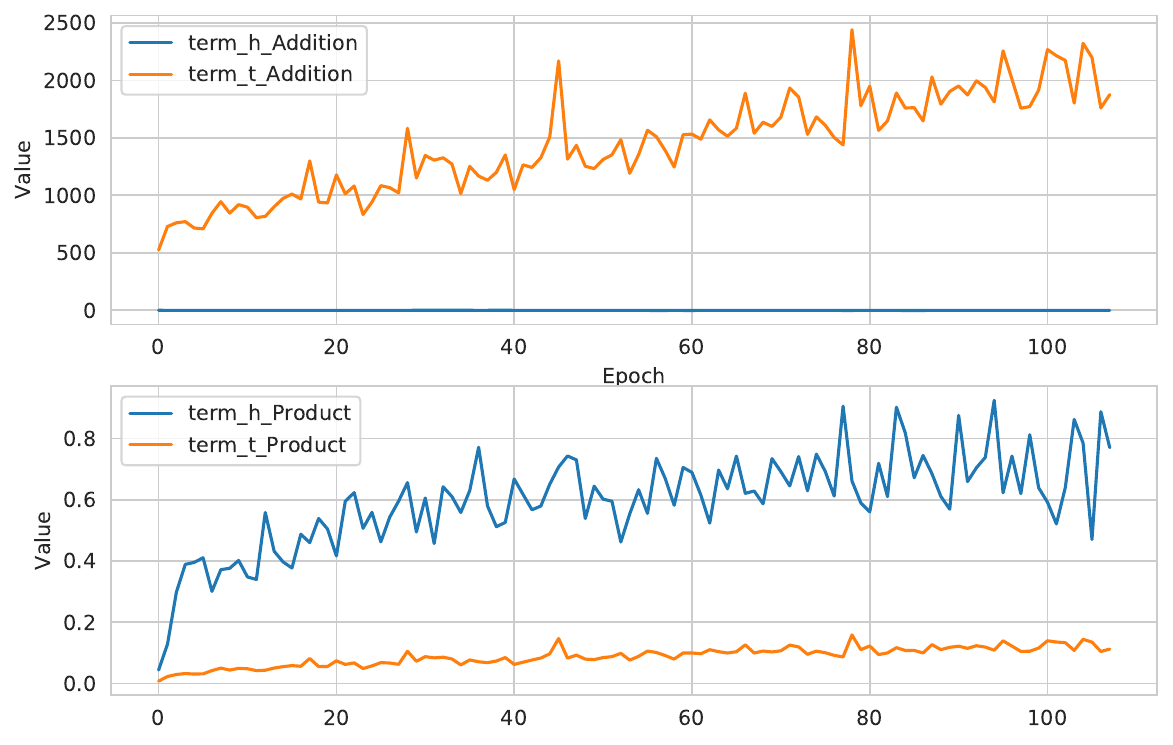}
\caption{Value Comparison During Training: Addition vs. Element-wise Product Operations. In addition operations, a significant magnitude disparity exists between the two values, with $\boldsymbol{h}$'s value approaching zero. Conversely, during element-wise products, $\boldsymbol{h}$'s value remains consistent, suggesting its role in scaling the prediction of future events based on past event information.
}
\label{value_comparison}
\end{figure}

\subsection{Ablation of Different Fusion Operations}
As discussed in the previous section, it is essential to visualize and compare the outputs from two distinct networks: one for the vector $\boldsymbol{h}$ and another for the time interval $\tau$ (refer to Figure \ref{whole}). We chose to conduct this comparative experiment on the Reddit dataset, as the differences are more pronounced with real-world data. The results are depicted in Figure \ref{value_comparison}, where the abscissa represents the epoch number and the ordinate indicates the output value. The orange and blue lines correspond to the outputs of $\boldsymbol{h}$ and $\tau$, respectively, processed through a simple network. The top and bottom sub-figures represent scenarios involving addition operations and element-wise product operations, respectively.
In the case of the addition operation, there is a significant disparity in the magnitude of the two values (the value corresponding to $\boldsymbol{h}$ is nearly 0). However, in the scenario with the element-wise product operation, the output value for $\boldsymbol{h}$ demonstrates fluctuations around 1. This observation suggests that the information encoded in past events has a scaling effect on the prediction of future events.

%% file: Content/Related_Work.tex
\section{Related Work}

\subsection{Neural Networks and TPPs}
Advances in Temporal Point Processes (TPPs) have significantly advanced through neural network integration. \cite{du2016recurrent} used a Recurrent Neural Network (RNN) to effectively compress historical events into vectors, key for intensity function formulation~\cite{li2018learning,huang2019recurrent,du2016recurrent,upadhyay2018deep}. Building on this, \cite{mei2017neural} introduced a Long-Short Term Memory (LSTM) network for complex intensity functions, requiring Monte Carlo for intractable likelihoods. Beyond neural networks, TPPs have been combined with techniques like noise-contrastive estimation~\cite{xiao2017wasserstein,yan2018improving,guo2018initiator,li2018learning,upadhyay2018deep}, offering training alternatives to maximum likelihood in TPPs.
Transformers, adapted for neural processes as in \cite{zhang2020self} and \cite{zuo2020transformer}, primarily focus on intensity function parameterization but face limitations, with advancements owing to their temporal modeling skills. \textbf{Our model, fundamentally orthogonal from transformer-based methods, concentrates on effectively modeling cumulative functions.} Acknowledging transformer-based techniques that incorporate meta-learning~\cite{bae2023meta} and contrastive learning~\cite{nguyen2022transformer}, our study uses an RNN as the main framework, setting fair and simple baselines for comparative analysis.

\subsection{Function Parameterization of TPPs}
Recent advancements in Temporal Point Processes (TPPs) have largely focused on developing more sophisticated methods for parameterizing intensity functions. The integration of mixtures of kernels, as expounded in the studies of \cite{taddy2012mixture}, \cite{tabibian2017distilling}, and \cite{okawa2019deep}, marks a significant advancement, providing increased flexibility in modeling temporal events. However, these kernel-based methods often encounter difficulties in encapsulating the complex, non-linear patterns frequently observed in real-world data. To tackle these intricacies, \cite{omi2019fully} introduced a methodology for directly computing the integral of intensity functions, aimed at a more comprehensive representation of temporal dynamics. Nonetheless, this approach occasionally fails to adhere to key TPP principles, such as ensuring the integral of the Probability Density Function (PDF) equals one, a constraint underscored in \cite{shchur2019intensity}'s work. In the evolving landscape of TPP research and development, our Cufun streamlines log-likelihood calculations. It leverages automatic differentiation to derive the density function from the Cumulative Distribution Function (CDF), \textbf{thus mitigating some computational instability} inherent in traditional intensity-based models.

\subsection{TPPs in Recommendation and Retrieval}
In recommendation systems and information retrieval, Temporal Point Processes (TPPs) are crucial, offering novel insights and significant solution enhancements. TPPs excel in retrieving continuous time event sequences and managing temporal data complexity for improved accuracy and relevance, as shown by \cite{gupta2022learning}. They are also adeptly used for graph-biased temporal dynamics in event propagation, enhancing network information understanding~\cite{wu2020modeling}. Additionally, TPPs in modeling non-Gaussian spatial-temporal processes~\cite{zheng2022modeling} provide a solid analytical framework. The use of knowledge- and time-aware modeling in sequential recommendation systems~\cite{wang2020make} and sequential hypergraphs for next-item recommendations~\cite{wang2020next} significantly advances prediction accuracy. These developments highlight TPPs' growing role in tackling challenges in recommendation and information retrieval. Our approach, CuFun, focuses on differential operations, aiming to boost performance and stability in neural network-integrated TPPs for recommendations and time series forecasting. Our proposed methodology, CuFun, which exclusively employs CDF functions, is poised to significantly elevate \textbf{the extreme/long-range events modeling and computational stability} in recommendation systems and information retrival.

%% file: Content/Conclusion.tex
\section{Conclusion}

This paper introduces the Cumulative Distribution Function-based Temporal Point Process (CuFun), a significant breakthrough in TPP modeling. CuFun's transition from conventional conditional intensity functions to CDFs effectively addresses key challenges in \textbf{computational stability, modeling of extreme or rare events, and capturing long-range dependencies.}
CuFun's implementation of a monotonic neural network for CDF estimation markedly improves its ability to incorporate historical data, thereby substantially increasing its predictive accuracy for future events. This advancement holds particular importance in domains requiring precise event sequence predictions, such as recommendation systems and information retrieval. Rigorous evaluations on various datasets, including both synthetic and real-world examples, have unequivocally established CuFun's superior performance compared to traditional function parameterizations.
Future research directions involve tailoring CuFun's architecture to specific use cases, expanding its integration with diverse data types, and assessing its applicability in other fields where accurate temporal event modeling is critical. The successful development of CuFun signals a promising trajectory in both theoretical and practical aspects of temporal data analysis, with particular implications for enhancing technologies in recommendation systems and information retrieval.
\section{Acknoledgement}
Special thanks to Zeyong Su for his generous and enthusiastic assistance and efforts. We are deeply grateful for his support.